\documentclass{article}

\usepackage[preprint]{neurips_rlbrew_mod_2024}

\usepackage[utf8]{inputenc} 
\usepackage[T1]{fontenc}    
\usepackage{hyperref}       
\usepackage{url}            
\usepackage{booktabs}       
\usepackage{amsfonts}       
\usepackage{nicefrac}       
\usepackage{microtype}      
\usepackage{xcolor}         
\usepackage{amsmath} 
\usepackage{graphicx}

\title{Tell me why: Training preference-based RL with human preferences and step-level explanations}

\author{
    Jakob Karalus  \\
    Ulm University \\
    Institute of Artificial Intelligence\\
    Ulm, Germany \\
     \texttt{jakob.karalus@uni-ulm.de} \\
}

\begin{document}

\maketitle

\begin{abstract}
Human-in-the-loop reinforcement learning allows the training of agents through various interfaces, even for non-expert humans. Recently, preference-based methods (PbRL), where the human has to give his preference over two trajectories, increased in popularity since they allow training in domains where more direct feedback is hard to formulate. However, the current PBRL methods have limitations and do not provide humans with an expressive interface for giving feedback. 
With this work, we propose a new preference-based learning method that provides humans with a more expressive interface to provide their preference over trajectories and a factual explanation (or annotation of why they have this preference). These explanations allow the human to explain what parts of the trajectory are most relevant for the preference. We allow the expression of the explanations over individual trajectory steps. 
We evaluate our method in various simulations using a simulated human oracle (with realistic restrictions), and our results show that our extended feedback can improve the speed of learning. 
\end{abstract}

\section{Introduction}

To allow for more flexible deployments of agents in all areas of human life, it is highly desirable that these can be quickly and easily adapted to different needs. While there are many different methods for training agents, Human-in-the-Loop Reinforcement Learning puts the responsibility on the human who gives feedback (or other methods like advice or demonstrations) from which the robots can learn (\cite{MoreInteractive}). This has the advantage that there is no need to program hard-coded actions or extensive modeling for planning or designing reward functions, which can be a cumbersome and brittle process in itself (\cite{RewardProblems}).
Pushing the responsibility directly to the human in the training process alleviates some of these problems while opening the learning process up to more non-technical users (or trainers). Naturally, there are multiple options for how a human can direct their feedback toward the agent. In this work, we want to focus on one of the more prominent methods of teaching agents by giving preferences over two trajectories, often called preferences-based reinforcement learning (PBRL). While Human-in-the-loop Reinforcement Learning (HRL) recently gained more interest from the research community, most research is focused on purely technical aspects with the same interface. We argue that extending that interface and giving humans more options to control learning is important.

Our work is also motivated by the concept of "scaffolding" (\cite{scaffolding}), which human teachers often use when teaching tasks to students. While scaffolding contains many techniques, like breaking the task into sub-parts, using verbal cues, or relating the content to other knowledge, we focus on a single technique in this work. Teachers often highlight essential features of the tasks (or solutions) to the student. This highlighting is often called factual explanations. 

Inspired by human scaffolding behavior, our main contribution is extending the common preference-based interface to allow the human to explain his decision more. More specifically, we allow the human to select timesteps of both trajectories (of the pair) to annotate which steps they deem important for his decision. This can be seen as a factual explanation. Our method then uses these explanations as additional training input. We archive this by generating an explanation of the reward model through gradients (e.g., saliency-based explanations) and comparing the generated explanations with the human explanations. This additional loss can be transparently integrated into the reward model learning of any preference-based reinforcement learning framework.

\section{Related Work}

Multiple works have attempted to extend the pipeline of Human-in-the-Loop RL to include more than (evaluative) feedback. 
\cite{Expand} implemented an HRL method that allows the user to highlight important areas of an image in addition to evaluative feedback. Our implementation can be used with any preference-based method (which can be used with any RL algorithm), while they designed a custom HRL method. Additionally, our proposed solution has the advantage that it is not restricted to an image-based state space but can deal with any state space. \\
\citet{reveal} showed that human explanation on individual feature space level can help to create better rewards models from an offline dataset of demonstrations. Again, their method is limited to cases where humans can understand the feature space. And secondly, their work is on offline learning, while we learn online. \\
\cite{FeatureQueries} allow the human to select which features influence their decision between two trajectories. While they show that their method can lead to increased learning performance, it is limited to a linear reward model, which limits its application quite drastically. Additionally, their work is limited in that they rely on complete trajectories (instead of segments in the case of PBRL).  From the human viewpoint, the work of \cite{FeatureQueries} is the closest comparison to ours. \\
\cite{Saran2020EfficientlyGI} improved the efficiency of Imitation Learning by collecting human gaze data, which is then used in an auxiliary loss constructed from the network attention and the human. Their idea of including additional information through an auxiliary loss, which is constructed through gradients of the network, is similar to the approach. But their focus is more on state regions, not timesteps, and the different settings of imitation learning. \\
\cite{CounterfactualTAMER} used counterfactuals (instead of factuals like our method) to increase the learning speed of TAMER (\cite{TAMER} alternative to preference-based RL for evaluative feedback). They reported increased learning performance, but their methods are for different types of human explanations and formulated in a different framework. \\
\cite{ITERS} lets users select part of a trajectory that they do not like. With this information, they perform reward shaping. While they allow humans to select parts of trajectories that they don't like, their annotations focus on either states or actions, not timesteps. Additionally, their setting of reward shaping still assumes the existence of an environmental reward and does not allow the training of agents purely from human feedback. \\
\cite{Wu2023FineGrainedHF} let the user select text parts (i.e., parts of their trajectory) to specify if that part is relevant for one of their pre-defined classes (Relevance, Factuality, etc..) While their work is solely focused on LLM/RLHF, the concept of selecting parts of the input trajectory is similar to our work. But this additional information is then used to train a pre-selected hard-coded set of reward models, which makes the transfer to other domains different. While our method is much more flexible and requires less oversight. 

Therefore, the opportunity to extend the setting to allow humans to give additional explanations has been investigated in different settings but not in the preference-based framework. Additionally, most of the work requires hand-crafted algorithms, while our contribution is much more independent of the concrete PBRL method. However, most methods reported increased learning speeds when including additional human annotations (in any form); therefore, it's realistic to expect changes in learning speed from our work.

\cite{RLHF-Blender} proposed an interface with multitudes of different annotation options for collecting human preferences over trajectories. Their focus was purely on the interface design, which is not connected to any real algorithm. To our knowledge, an explanatory annotation of timesteps inside trajectories is impossible in their theoretical interface because they drew heavy inspiration from existing research. This shows the importance of showcasing the technical feasibility of possible feedback interface avenues so they can be included in future human-computer-interaction research. \cite{RLHFSurvey} specified "fine-grained feedback" as a possibility for problems where precise information is necessary to solve the task.

In settings other than PBRL, the concept of using gradient-based explanations to enhance training has been applied successfully. 
Saliency-based methods have been used in other contexts to speed up the learning process or force the learner to learn the right reasons. Common examples where gradient-based explanations have been used (in supervised learning) to accelerate learning are ExpectedGradients (\cite{ExpectedGradients}), Right-for-the-right-reasons (\cite{rightfortherightreasons} or \cite{saliencytraining}. These ideas are not yet been explored in PBRL. 

\section{Background}

We consider a setting wherein the sequential interactions between an agent and its environment are formally characterized by a Markov Decision Process (MDP), a mathematical framework comprising states, actions, and transitions. Within this formalism, the system evolves discretely at time intervals \(t.\) 
Within the episodic framework, the agent's involvement persists until reaching a terminal time step \(T\), signifying the conclusion of an episode. This temporal progression is succinctly represented by the trajectory, expressed as the ordered sequence \((s_1, a_1), \ldots, (s_T, a_T),\) encapsulating the observation-action pairs throughout an episode.

In RL the agent experiences rewards at each time step. In our Human-in-the-Loop context, we abstain from presuming access to such rewards. Instead, we introduce a human overseer controlling the agent's task intention through two distinct feedback channels. First, their preferences (as generally in PBRL) and second, their explanations (our contribution).

\subsection{Preference-based Reinforcement Learning}
The main goal of preference-based reinforcement learning (PBRL, \cite{Christiano2017DeepRL}) is to learn a reward function, denoted as $\hat{r}_\psi$, from a set of expressed segment pairs. These pairs of segments are collected from the human throughout the training process. 
Within this framework, a segment, \(\sigma\), is defined as a sequence of states and actions, \(\{s_k, a_k, ..., s_{k+H}, a_{k+H}\}\). Preferences, denoted as \(y\), are elicited for segments \(\sigma_0\) and \(\sigma_1\), with \(y\) representing a distribution indicating the preferred segment, i.e., \(y \in \{(0, 1), (1, 0), (0.5, 0.5)\}\). This evaluative judgment is recorded in a dataset \(D\) as a triple \((\sigma_0, \sigma_1, y)\). This dataset of binary preference can be then used to train a reward model with the Bradley-Terry model (\cite{BradleyTerry}):

\begin{equation}
    P_\psi[\sigma_1 \succ \sigma_0] = \frac{\exp \sum_t \hat{r}_\psi(s_{1t}, a_{1t})}{\sum_{i \in \{0,1\}} \exp \sum_t \hat{r}_\psi(s_{it}, a_{it})}
\end{equation}

Here, \(\sigma_i \succ \sigma_j\) denotes that segment \(i\) is preferred to segment \(j\). This formulation states that the probability of preferring a segment exponentially depends on the reward function's sum over the segment. While \(\hat{r}_{\psi}\)  itself is not inherently a binary classifier, the learning procedure is akin to binary classification, where a supervisor supplies labels \(y\). To learn the reward function, instantiated as a neural network with parameters \(\psi\), the following loss is minimized:

\begin{equation}
    \label{eq:preference_loss}
    \mathcal{L}_{\text{preference}} = - \mathbb{E}_{(\sigma_0, \sigma_1, y) \sim \mathcal{D}} \left[ y_{0} \log P_\psi[\sigma_0 \succ \sigma_1] + y_{1} \log P_\psi[\sigma_1 \succ \sigma_0] \right]
\end{equation}

Even in PBRL the final goal is to learn a policy \(\pi\) that maximizes the expected cumulative reward. The reward function  \(\hat{r}_{\psi}\) obtained from preferences is a surrogate to the true, unknown reward function. Therefore \(\hat{r}_{\psi}\) is utilized instead of the actual reward signal while optimizing the policy. The policy can be learned through standard RL algorithms like PPO (\cite{Schulman2017ProximalPO}) or SAC (\cite{Haarnoja2018SAC}).

\subsection{Training Guided by Saliency-Based Methods}
Saliency-based explanation methods allow an understanding of neural network decision-making processes, offering a quantitative means to assess the impact of input features on model predictions. Consider a neural network \(f_\theta\) with parameters \(\theta\), trained on a dataset \(\mathcal{D} = \{(x_i, y_i)\}_{i=1}^N\), where \(x_i\) denotes input samples and \(y_i\) represents corresponding labels. To calculate a saliency-based explanation (or saliency map) for an input $x_i$, a normal loss between the prediction of the network $f_\theta$ and the label $y_i$ is calculated. That loss then allows us to calculate the gradients of the input $x_i$ (with respect to the before-mentioned loss). These gradients are then used as an explanation of the prediction.

One notable saliency-based method is SmoothGrad (\cite{smoothgrad}), designed to stabilize saliency maps. The smoothed saliency map \(S_{\text{SmoothGrad}}\) is computed as the average gradient over \(N_{\text{smooth}}\) perturbed samples:
\begin{equation}
    \label{eq:smooth_grad}
    S_{\text{SmoothGrad}} = \frac{1}{N_{\text{smooth}}} \sum_{i=1}^{N_{\text{smooth}}} \nabla_{x} f_\theta(x + \epsilon_i)
\end{equation}

Here, \(\epsilon_i\) represents the perturbation, and \(\nabla_{x} f_\theta(x + \epsilon_i)\) is the gradient of the model's prediction with respect to the perturbed input. This approach introduces controlled noise to the input, providing a more stable saliency map. The resulting map highlights the regions crucial for the network's predictions, contributing to a nuanced understanding of feature importance in neural network decision-making. Similar methods like IntegratedGradients \cite{integratedgradients} calculate multiple explanations with respect to a baseline and average the results to increase stability.

Saliency-based training methods (\cite{saliencytraining}), such as ExpectedGradients (\cite{ExpectedGradients}) and Right-for-the-Right-Reasons (\cite{rightfortherightreasons}), allow the shaping of the training dynamics of neural networks. These approaches utilize saliency information to refine the training process, enhancing model interpretability and predictive performance.
Right-for-the-Right-Reasons emphasize salient features contributing to correct predictions during training, steering the model towards more meaningful representations. This is done by adding additional terms to the loss term. Instead of a single loss term between the prediction's distance from the target (i.e., being right), a second (weighted) term is added to constrain the distance between a saliency-based explanation and a given explanation (i.e., the right reason). 

ExpectedGradients \cite{ExpectedGradients} allows to guide the training even in cases where no ground-truth explanations are available. This is done by imposing priors on the structure of the explanation. For example, explanations should be either locally-smooth or sparse regarding features.
These saliency-guided training strategies embody a synergy between interpretability and maintaining predictive accuracy.

\section{Implementation}

\begin{figure}
    \centering
    \includegraphics[width=\columnwidth]{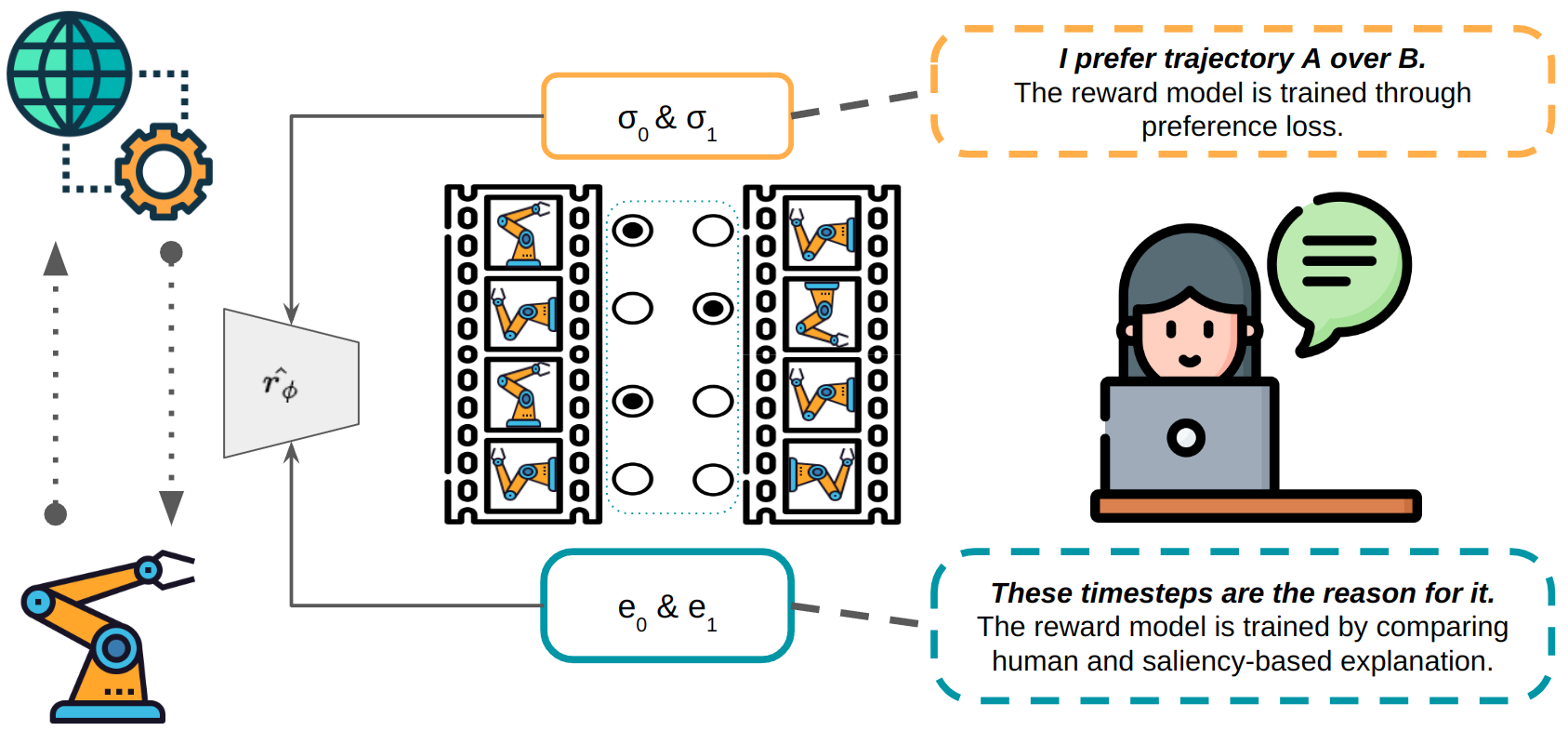}
    \caption{Overview of our approach. Left: The agent optimizes his policy with respect to the trained proxy reward model. Right: The proxy reward model is trained from human preferences over trajectories and the importance of each timestep in the trajectory. Humans provide both.}
    \label{fig:approach}
\end{figure}

Our main contribution is the extension of the feedback possibilities in PBRL for humans. In addition to their preferences between the two trajectories, the human can now explain which timesteps they deems essential for their decision. This evaluative judgment, accompanied by detailed binary explanations provided by the human over timesteps, is recorded the dataset \(\mathcal{D}\) as a quintuple \((\sigma_0, \sigma_1, y, \text{e}_0, \text{e}_1)\). Here, \(\text{e}_0\) and \(\text{e}_1\) directly represent binary vectors of human-provided explanations, each having the same length as the corresponding segments \(\sigma_0\) and \(\sigma_1\). These binary vectors, \(\text{e}_0\) and \(\text{e}_1\), serve to elucidate the significance of individual timesteps within their respective segments, providing binary indications (0 or 1) of the human's perceived importance at each step. 
Figure \ref{fig:approach} shows an overview of our approach. 

\subsection{Reward Learning with annotated preferences} 
Incorporating preference-based learning alongside annotations necessitates introducing two additional steps within the learning process. First, we must predict an explanation $\hat{e}$ and then compare this generated explanation with the human explanation $e$.

Initially, an explanation of the existing prediction (of the reward model) concerning the current trajectory pair is formulated. The generation of this explanation leverages established saliency-based methods, and we employ the SmoothGrad technique (see Equation~\ref{eq:smooth_grad}) in our implementation. Here, a set of \(N_{\text{smooth}}\) perturbations is generated for a given input trajectory by sampling from a normal distribution with mean and standard deviation parameters derived from the input trajectory. On these perturbations, we then calculate the saliency-based explanations. 

\begin{equation}
    \xi(\sigma) = \left[ \sum \left| \frac{1}{N_{\text{smooth}}} \sum_{i}^{N_{\text{smooth}}} \nabla_{s,a} \hat{r}_\psi(s + \epsilon_i, a + \epsilon_i) \right| \: \right] \: \forall s,a \in \sigma
\end{equation}

$\epsilon$ represents the perturbation of the input. We then predict the outcome (reward) with the reward model $\hat{r}_\psi$ and calculate the gradient with respect to the input, i.e., the state and action. This done for all timesteps $t$ in the segment $\sigma$.

Since the human explanations (\(\text{e}_1\) and \(\text{e}_2\)) are binary vectors with the length of the segments \((\sigma_0, \sigma_1) \), but the saliency-based explanations $\hat{e}$ are of shape $(s,a)_t$ (a concatenation of state and actions space for each timestep in the segment). Therefore, the generated explanations must be transformed from the state \& action frame space to a step-level explanation. To achieve this, we sum the absolute values over each timestep to transform the frame-based values into a step-level explanation. These unnormalized values then serve as logits for the binary-cross entropy loss, aligning the calculated explanation with the provided annotation, the latter specified by the user's annotations.

The annotation  loss \(L_{\text{annotation}}\) is the standard binary multi-class cross-entropy between our explanation-logits and the true explanation:
\begin{equation}
    \label{eq:annotation_loss}
    L_{\text{annotation}} =  \frac{1}{2N} \sum_{e, \sigma \in \{ (e_0, \sigma_0), (e_1, \sigma_1) \} \in \mathcal{D}} [ -(e \cdot \log (f(\xi(\sigma)))) + (1 - e \cdot \log (1 - f(\xi(\sigma))))]
\end{equation}

where $e$ is the explanation of the user (for the human preference $y$) and $\xi(\sigma)$ generates the explanation on a trajectory step level. 
$N$ is the total number of preferences (i.e., a pair of segments) in the database $D$. The user's preferences are $e$, and our generated preferences are transformed with $\xi(\sigma)$. Since  $\xi$ produces normalized logits, we apply the sigmoid function for $f$. To relax the learning constraints, we employ label smoothing for the human explanations $e$ to relax the constraints of having the correct explanations.

Theoretically, we could allow the user to give specific annotations, i.e., highlighting single entries in the state. Still, there are reasons to limit the annotations' granularity to the trajectory's timestep level. First, the timestep level of the trajectory is quite application-independent and does not require the human to fully understand lower levels (like the state and action space). For example, in many robots, the state space includes a mix of different sensors (LIDAR, Camera, etc.), which can be hard to understand fully for humans. Lower levels also drastically increase the mental load on humans due to the massive number of options. Therefore, we want to show in this work that even explanations on the timestep level contain enough information to increase learning speed.

\subsection{Structural Loss}

Like ExpectedGradients, we want to constrain the structure of our generated explanation and impose a structural loss on the derived explanation. Since most human explanations are sparse, we apply this prior to our generated explanation and force the generated explanations to reflect this sparse characteristic. This structural loss is realized by applying an L1 norm to the calculated explanation.

Mathematically, the structural loss \(L_{\text{structural}}\) is formulated as:
\begin{equation}
    \label{eq:structural_loss}
     L_{\text{structural}} = \frac{1}{2N} \sum_{\sigma \in \{ \sigma_0, \sigma_1 \} \in \mathcal{D}}  \|\xi(\sigma)\|_{1}
\end{equation}

where $N$ denotes the total number of trajectories in $ D$, and $\xi$ generates a step level explanation for the segment $\sigma$. Incorporating this structural loss contributes to the regularization of the explanation, fostering a more coherent and interpretable representation.

To derive our reward model \(\hat{r}_{\phi}\), we integrate all three distinctive loss terms: the primary preference loss (Equation~\ref{eq:preference_loss}), the annotation loss (Equation~\ref{eq:annotation_loss}), and the structural loss (Equation~\ref{eq:structural_loss}), merging them into a unified loss term. The incorporation of the additional losses is accompanied by weightings (\(\alpha_1\) and \(\alpha_2\)):

\begin{equation}
    L_{\text{total}} = L_{\text{preference}} + \alpha_1 \cdot L_{\text{annotation}} + \alpha_2 \cdot L_{\text{structural}}
\end{equation}

This loss formulation ensures the simultaneous consideration of preference-based learning, explanation fidelity, and structural regularization in the training of the final reward model $\hat{r_{\phi}}$.

\subsection{Policy Learning}
To learn an actual policy from the learned reward function, we use the same setup as PEBBLE \cite{2021pebble}, in which the offline learning algorithm SAC is leveraged. Like in PEBBLE, we update the replay buffer every time the reward function is updated. This allows the agent to always learn from the latest experience. Since we extend from PEBBLE, it's also the choice of our baseline for the evaluation.

\section{Evaluation}
\label{sec:evaluation}
We evaluate our changes in three common robotics environments, Walker, HalfCheetah, and Hopper of the mujoco suite. Most of the hyperparameters are taken from our baseline PEBBLE (\cite{2021pebble}), which we then briefly tuned only on the baseline (since we argue that our work should be viewed as an extension to the existing method, we believe it should not deserve a full hyperparameter search). The ballpark for the hyperparameter of our extension (i.e., the weight of additional loss terms) was also taken from similar methods (\cite{rightfortherightreasons}) and briefly sanity-checked. The agent trains for 2 million steps in each environment, with a total feedback budget of 700 comparisons. A full overview of all hyperparameters can be found in Appendix \ref{sec:appendixhparam}

While we are learning in an HRL setting, where we don't have access to the true reward function, we still use the true reward function for the evaluation. To provide a solid measurement of the learning progress, we perform every 10k environment steps performance measurement of the agent. This measurement consists of 5 episodes in a newly seeded (with a different random seed than the current iteration) environment. At each measurement, we collect total true environmental rewards and calculate the mean between the 5 measurement episodes. With this setting, we collect 200 measurements throughout the whole training process (of a single evaluation run). This extensive collection of measurements gives us a more stable estimation of the true performance (of the agent) throughout training than the normally used rollout mean rewards. These collected measurements are then aggregated and normalized between zero and one. We use the same procedure proposed by \citet{rliable} to calculate mean scores and confidence intervals via bootstrapping over 10 runs in each condition.

We opted to create a synthetic oracle instead of actual humans to allow for an evaluation with enough statistical significance. For the comparison, we used the environmental ground truth rewards (since in DM-control, it exists here); for the trajectory step-level explanation, we resorted to an XAI technique called IntegratedGradients (\cite{integratedgradients}) to extract the explanations from a pre-trained policy. 

The random seed is separated between the environment dynamics and the network/agent dynamics. We select the same starting seeds for the comparisons between our work and the baseline (to reduce the impact of "favorable" seeds).

Since the training with a perfect oracle is not a realistic setting, we test the stability of our extension properly and use the five irrationalities of human teachers as defined by the standard B-Pref benchmark (see \cite{BPref}). These different common failure patterns each represent different "sub-optimality" that can occur in human feedback. We also train with the perfect case (the oracle case).
\begin{figure}
    \centering
    \includegraphics[width=\columnwidth]{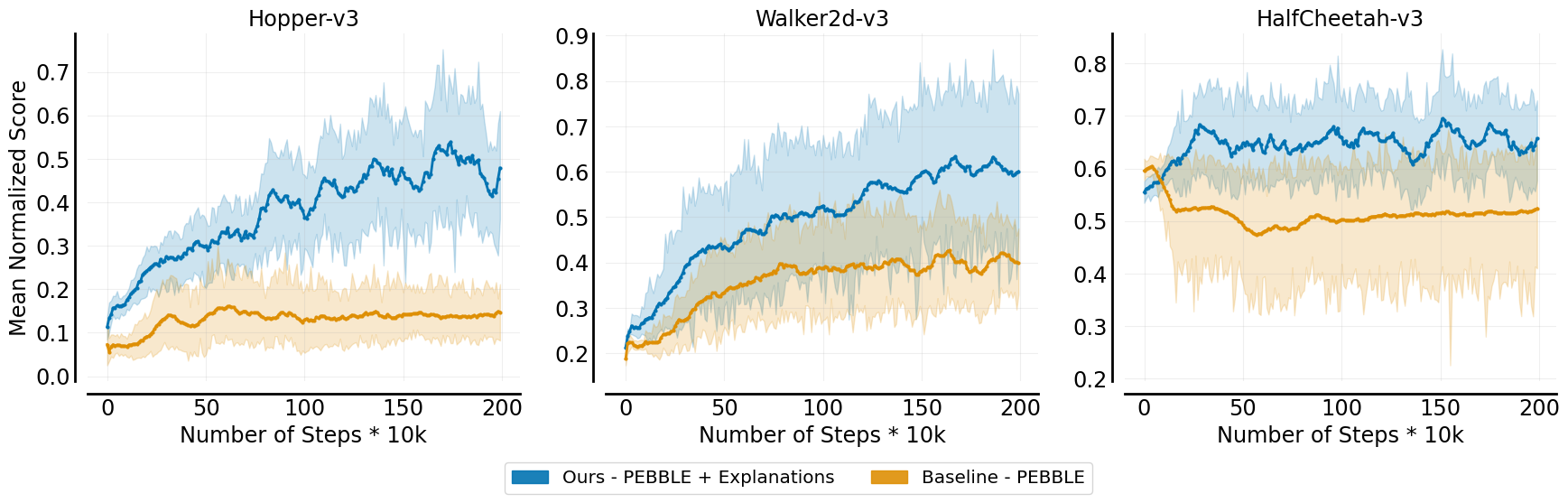}
    \caption{The mean throughout the training process in each environment. Shaded areas represent the confidence interval. Higher is better. }
    \label{fig:training_progress}
\end{figure}

\subsection{Results}

As visible in Figure \ref{fig:training_progress}, we can achieve better mean rewards (the mean of 10 runs, each point is measured in 5 evaluation episodes, as described in section \ref{sec:evaluation}) results than the baseline with our extension in all environments with the perfect oracle. While in the Walker and HalfCheetah environments, the mean (bold line) is higher, the confidence interval (shaded areas) still overlaps; therefore, we can't extensively say that our methods are conclusively better. Only in the Hopper case can we conclusively answer that the human explanation helps the reward model learn faster. While not pictured, both methods converge to the same reward in the long run. However, our method's major motivation and driving force was increasing learning speed, especially in the earlier phases when the human is still in the loop. 

\begin{figure}
    \centering
    \includegraphics[width=\columnwidth]{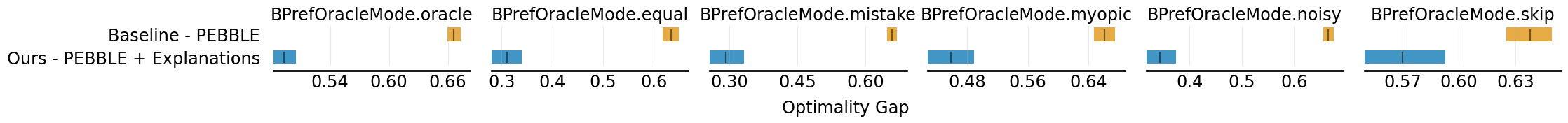}
    \caption{Optimality Gap (how fast a run converges to the ideal score, lower is better) under different human irrationality. The vertical line visualizes the mean optimality gap, and the bars represent the confidence interval.}
    \label{fig:optimality_gap}
\end{figure}

Since we want to highlight the increase in learning speed, we use the optimality gap as our primary metric for further evaluation since it condenses the learning speed into a single, more understandable metric. In Figure \ref{fig:optimality_gap}, we use the optimality gap to compare our modifications in the different irrationalities defined by BPref. Notable is the significance (indicated by no overlapping bars, as defined in \cite{rliable}) between our work and the baseline in quite a few cases. 
These results confirm that our extension leads to better learning performance, even in settings where the human feedback follows a more realistic pattern.

\section{Discussion}


\textbf{Additional mental load} Labelling trajectories not only with the binary preference but also with the annotation of which parts are better/worse does, of course, require more mental effort than pure preference-based (or other evaluative approaches). In contrast, we showed that these annotations speed up the learning process, and therefore, less total human interaction is needed. We believe this trade-off has to be evaluative on a case-by-case basis since it's dependent on multiple factors like the environment, the learning algorithm, the feedback interface, and the expertise and type of trainers. Additionally, we want to highlight that our annotations might relieve some of the frustrations human trainers often have when using a very restricted feedback interface (binary in the case of preference-based learning). Our methods give the human trainer more avenues to express his feedback, which should increase user experience.

\textbf{Granularity of the annotations} While it's technically feasible to annotate whole states/actions in a trajectory and annotate specific parts of the state space directly, we refrain from evaluating this scenario. We believe preference-based learning shines in settings where the human can't highlight single points in the state-space because of the complexity of the state space. We believe in areas where its possible for human to understand single points in the state-space, methods like TAMER \cite{TAMER} are probably a better alternative.

\textbf{Evaluation with a simulated oracle}
Throughout this work, we evaluated our modification with a synthetic oracle instead of an actual human. While this is a weakness of the evaluation, it also allowed us to collect more runs (with different random seeds) and perform different ablations. This increase makes the statistical analysis more sound. We believe this benefits the research community more than a human study with a few participants, which would not allow us to show our gains with (statistical) confidence.

\section*{Conclusion}
With this work, we showed how preference-based reinforcement learning could be extended so humans can give not only preferences but also explanations for their decisions. 
In experiments, we showed that with these explanations, we can increase the performance of current state-of-the-art PBRL even further. Not only in the ideal case but also under a variety of different human irrationalities, our extension either increases the performance or does not reduce it. 

Our method, as presented here, is meant as a first demonstration, favoring simplicity over sophistication. A significant positive attribute of PBRL is that its RL part is mainly decoupled from the human-in-the-loop, which means that future advantages in RL, which bring new algorithms, can be easily included. Our extensions continue that attribute since we only require changes to the reward model, not the agent. 

Overall, we showed potential in opening up the interface in human-in-the-loop/preference-based RL to include more natural and human-like feedback mechanisms instead of relying only on binary feedback.

\bibliography{main}
\bibliographystyle{plainnat}

\appendix

\section{Hyperparameters}
\label{sec:appendixhparam}
We use the following hyperparameters for our experiments. Most of the common hyperparameters are from \cite{2021pebble} or \cite{Christiano2017DeepRL}.
\begin{table}[htbp]
    \begin{center}
        \begin{tabular}{ll}
            \multicolumn{1}{l}{\bf Hyperparameter}  &\multicolumn{1}{l}{\bf Value}
            \\ \hline \\
            \multicolumn{2}{c}{\bf SAC} \\
            \\ \hline \\
            Learning rate          &0.0003 \\
            Batch size        & 512 \\
            Buffer size & 1000000 \\
            $\tau$ & 0.005 \\
            $\gamma$ & 0.99 \\
            \\ \hline \\
            \multicolumn{2}{c}{\bf Reward Model} \\
            \\ \hline \\
            Ensemble size & 3 \\
            Number of hidden units & 300 \\
            Number of hidden layers & 3 \\
            Learning rate & 5e-4 \\
            Batch size & 32 \\
            Adam $\beta's$ & (0.9, 0.9) \\
            AdamW weight decay & 0.05 \\

            \\ \hline \\
            \multicolumn{2}{c}{\bf Our} \\
            \\ \hline \\
            $\alpha_1$ & 0.25 \\
            $\alpha_2$ & 0.1 \\    
            Smoothgrad number of pertubations & 16 \\
            Smoothgrad noise & 0.01  \\
            \\ \hline \\
            \multicolumn{2}{c}{\bf Feedback} \\
            \\ \hline \\
            Maximum number of feedback & 700 \\
            Feedback per interval & 70 \\
            Feedback is given every X steps & 20000 \\
            Maximum segment size & 50 \\
            Segment sampling scheme & Disagrement \\
        \end{tabular}
    \end{center}
    \caption{Hyperparameters}
    \label{tab:exampleTable}
\end{table}

\subsection{Compute Power}
Most of the experiments were performed on a single NVIDIA A100 with a single seed run in environments that took roughly 4 hours. Many runs were calculated in parallel on the same machine to increase speed.

\end{document}